\documentclass[10pt,twocolumn,letterpaper]{article}

\usepackage{iccv}
\usepackage{times}
\usepackage{epsfig}
\usepackage{graphicx}
\usepackage{amsmath}
\usepackage{amssymb}

\makeatletter
\@namedef{ver@everyshi.sty}{}
\makeatother

\usepackage{pgfplots}
\pgfplotsset{compat=1.12}
\usepackage{tabularx}
\usepackage{censor}
\usepackage{subcaption}
\usepackage{multirow}

\usepackage[pagebackref=true,breaklinks=true,letterpaper=true,colorlinks,bookmarks=false]{hyperref}

\iccvfinalcopy 

\def\iccvPaperID{3502} 

\ificcvfinal\fi

\newcommand{\Loss}[0]{{\mathcal{L}}}

\begin{document}
\title{CityFlow-NL: Tracking and Retrieval of Vehicles at City Scale \\ by Natural Language Descriptions}

\author{Qi Feng\\
Boston University\\
{\tt\small fung@bu.edu}
\and
Vitaly Ablavsky\\
University of Washington\\
{\tt\small vxa@uw.edu}
\and
Stan Sclaroff\\
Boston Univeristy\\
{\tt\small sclaroff@bu.edu}
}

\newcommand{\fred}[1]{\textcolor{red}{#1}}

\newcommand{\remove}[1]{\textcolor{red}{\textbf{Consider Remove:} #1}}

\definecolor{C1}{RGB}{252,202,108}
\definecolor{C2}{RGB}{200,91,108}
\definecolor{C3}{RGB}{49,54,88}
\definecolor{C4}{RGB}{242,163,94}
\newcommand{\rulesep}{\unskip\ \vrule\ }

\pgfplotscreateplotcyclelist{default}{
  smooth, color=red, line width=0.60mm \\
  smooth, color=blue, line width=0.40mm \\
  smooth, color=cyan, line width=0.40mm \\
  smooth, color=magenta, line width=0.40mm \\
  smooth, color=violet, line width=0.40mm \\
  smooth, color=C1, line width=0.40mm \\
  smooth, color=C2, line width=0.40mm \\
  smooth, color=C3, line width=0.40mm \\
  smooth, color=C4, line width=0.40mm \\
}

\maketitle

\begin{abstract}
  Natural Language (NL) descriptions can be one of the most convenient or the
  only way to interact with systems built to understand and detect city scale
  traffic patterns and vehicle-related events. In this paper, we extend the
  widely adopted CityFlow Benchmark with NL descriptions for vehicle targets and
  introduce the CityFlow-NL Benchmark. The CityFlow-NL contains more than 5,000
  unique and precise NL descriptions of vehicle targets, making it the first
  multi-target multi-camera tracking with NL descriptions dataset to our
  knowledge. Moreover, the dataset facilitates research at the intersection of
  multi-object tracking, retrieval by NL descriptions, and temporal localization
  of events. In this paper, we focus on two foundational tasks: the Vehicle
  Retrieval by NL task and the Vehicle Tracking by NL task, which take advantage
  of the proposed CityFlow-NL benchmark and provide a strong basis for future
  research on the multi-target multi-camera tracking by NL description task.
\end{abstract}

\section{Introduction}
Understanding city-wide traffic patterns and detecting specific vehicle-related
events in real time has applications ranging from urban planning and
traffic-engineering to law enforcement. In such applications, Natural Language
(NL) descriptions can be one of the most convenient or the only way to interact
with a computer system, \eg taking as input an informal description provided by
a bystander. Thus, it becomes necessary to detect, track and retrieve vehicles,
potentially seen from multiple spatio-temporally disjoint viewpoints, based on
NL queries.

\begin{figure}
    \resizebox{\columnwidth}{!}{
      \includegraphics[width=\textwidth, page=1, trim=10cm 4cm 8cm 4.2cm, clip]{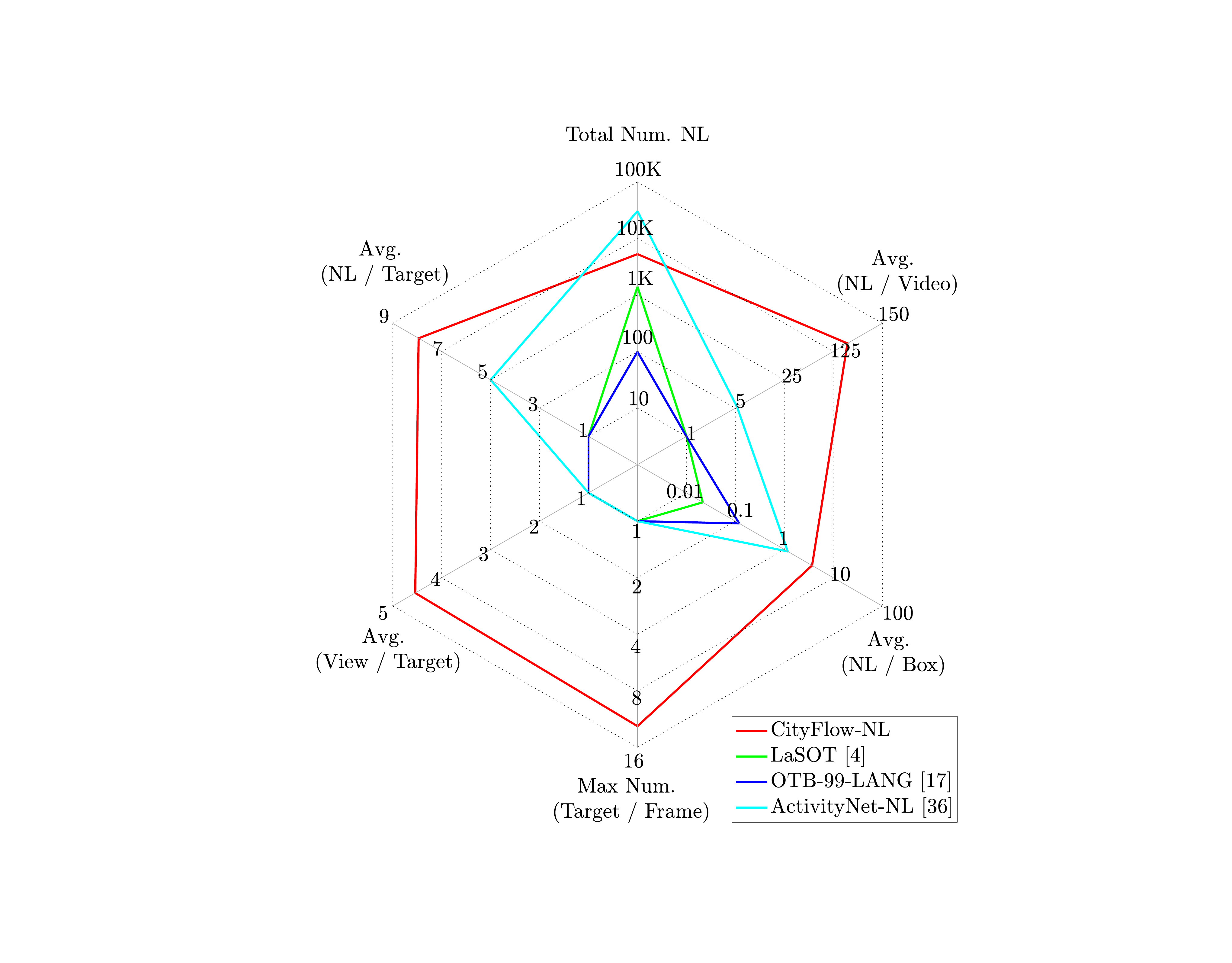}
    }
    \caption{
    Comparison between NL annotated visual tracking datasets. The proposed
    CityFlow-NL is the first natural language multi-view multi-target tracking
    benchmark.
    }
    \label{fig-star}  
    \vspace{-0.2 in}
  \end{figure}
  
\begin{figure*}
  \includegraphics[width=\textwidth, page=2, trim=0 14cm 0 0, clip]{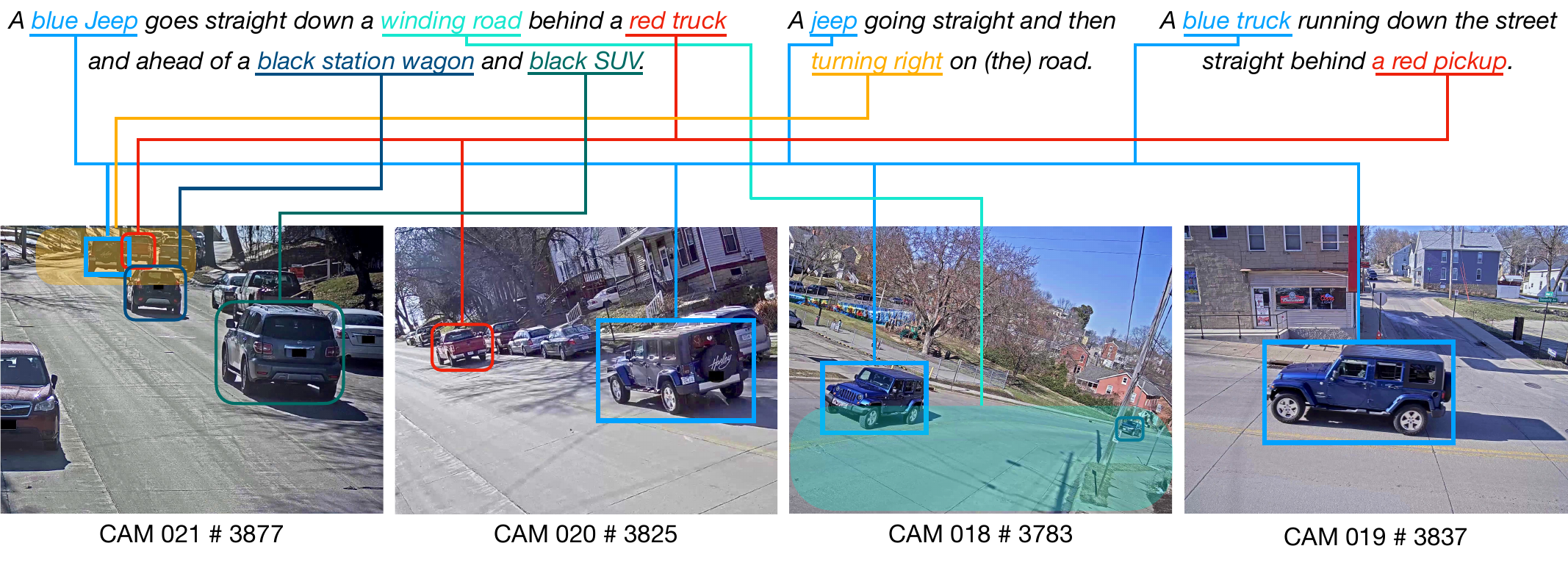}
  \caption{
    Example frames and NL descriptions from the proposed CityFlow-NL dataset.
    Crowdsourcing workers annotate the target vehicle using a carefully designed
    multi-camera annotation platform (Fig.~\ref{fig-website}). NL descriptions we collect tend to describe vehicle color/type
    (\eg \textit{blue Jeep}), vehicle motion (\eg \textit{turning right} and
    \textit{straight}), traffic scene (\eg \textit{winding road}), and relations
    with other vehicles (\eg \textit{red truck, black SUV}, \etc).
  }
  \label{fig-nl-annotations}
  \vspace{-0.2 in}
\end{figure*}

Motivated by this demand, we define and address the task of Multi-Target
Multi-Camera (MTMC) by NL Description. The goal of this task is to track a
specific target described by an NL query within a sequence of multi-view video
in the form of a sequence of bounding boxes on each camera view. While it shares
similarities with other Vision+NL tasks like single object tracking (potentially
without distraction) by NL~\cite{feng2020real,feng2019robust,li2017tracking},
NL-based video retrieval~\cite{anne2017localizing,zhang2019man}, and
spatio-temporal localization by
NL~\cite{gavrilyuk2018actor,hu2016segmentation,yamaguchi2017spatio}, the
proposed task requires both temporal and spatial localization of the target from
a multi-view video via an NL query. Up until now, an added challenge in
developing and evaluating algorithms for the proposed task is the lack of
realistic datasets.

Therefore, we introduce CityFlow-NL, an open dataset designed to facilitate
research at the intersection of multi-object tracking, retrieval by NL
specification, and temporal localization of events. Our benchmark is derived
from CityFlow~\cite{tang2019cityflow}, which is itself a public dataset that has
been at the center of several recent workshop-challenges focused on MTMC
tracking and
re-identification~\cite{naphade2017nvidia,naphade20182018,naphade20192019,naphade20204th}.
As a result, the proposed CityFlow-NL dataset exhibits a diversity of real-world
parameters, including traffic density and patterns, viewpoints, ambient
conditions, \etc.

The NL descriptions are provided by at least three crowdsourcing workers, thus
capturing realistic variations and ambiguities that one could expect in such
application domains. Crowdsourcing workers describe the target vehicle using a
carefully designed multi-view annotation platform. The NL descriptions tend to
describe the vehicle color, the vehicle maneuver, the traffic scene, and
relations with other vehicles. Example NL descriptions and vehicle targets are
shown in Fig.~\ref{fig-nl-annotations}.

Although the motivation for building the CityFlow-NL dataset is to tackle the
real-world problem of MTMC by NL, the introduction of the CityFlow-NL also opens
up research possibilities for the retrieval/tracking by NL tasks to more
advanced topics, \eg action recognition by NL and spatio-temporal relation
understanding problems. In this paper, we propose two basic tasks: the vehicle
retrieval by NL description task and the vehicle tracking by NL description
task. We consider these tasks as a foundation for more advanced Vision+NL tasks,
including the MTMC by NL task. 

Given a sequence of a \emph{single-view} video and an NL description that
uniquely describes a vehicle target in the video, the goal of the vehicle
tracking task is to predict a spatio-temporal localization in the form of a
sequence of bounding boxes for the target. The vehicle retrieval task is a
simpler form of the tracking task, in which the spatial localizations of all
candidate vehicles are given and the goal is to retrieve the target specified by
the NL query. These tasks are related to object tracking and retrieval systems
and can be evaluated in a similar way, but the NL query setup poses unique
challenges. For example, relationship expressions between targets
(Fig.~\ref{fig-nl-annotations}) and spatial referring expressions
(Fig.~\ref{fig-vtn-results}) have not been investigated in visual trackers.

\section{Related Works}
\label{sec-related-works}
In the past decade, researchers have started to look into exploiting natural
language understanding in computer vision tasks. These models usually
consist of two components: a language model and an appearance model to learn a
new feature space that is shared between both NL and visual
appearance~\cite{johnson2016densecap,vinyals2015show}.  More recent object
detection and vision grounding models~\cite{hong2019learning,yang2019fast}
jointly exploit vision and NL using Siamese networks and depth-wise
convolutional neural networks.

The VisualGenome~\cite{krishna2017visual}, Flickr-30K~\cite{young2014image} and
Flickr-30K entities~\cite{plummer2015flickr30k} benchmarks facilitate research
in understanding language grounded in visual contents. These image datasets
provide detailed descriptions of regions in an image but still lack temporal
information that can enable systems to better handle the temporal context and
motion patterns in the NL descriptions.

OTB-99-LANG~\cite{li2017tracking} is the first open dataset that provides NL
descriptions for single object tracking, followed by LaSOT~\cite{fan2019lasot}
which is a large scale single object tracking benchmark annotated with NL
descriptions. These datasets enable researchers to perform spatial localization
of targets based on NL descriptions of the target objects. However, both of
these datasets are annotated with one NL description for the target object for
the entire sequence. The NL descriptions in OTB-99-LANG and LaSOT is limited to
a sentence or a phrase during annotation and can be ambiguous when used for
tracking. These NL descriptions are not able to describe the motion pattern of
the target~\cite{feng2020real}, especially over a longer time interval.

\begin{figure}
  \resizebox{\columnwidth}{!}{
    \input{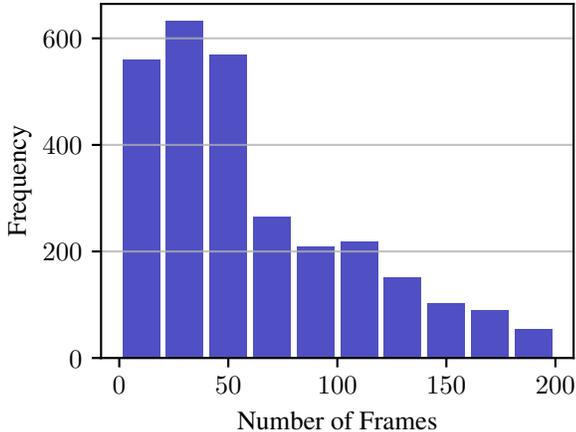}
  }
  \caption{
  Histogram of number of frames vehicle targets shows in camera. The average
  number of frames is 75.85 frames, during which a vehicle target typically
  conducts a single maneuver and can be precisely described by an NL description
  in one sentence. Targets in existing single object tracking with
  NL benchmarks, \eg OTB-99-LANG~\cite{li2017tracking} and
  LaSOT~\cite{fan2019lasot}, are not or cannot be precisely described by a
  sentence due to the length of the videos in these datasets.
  }
  \label{fig-distribution}
  \vspace{-0.2 in}
\end{figure}

On the other hand, temporal localization of targets in videos using NL
descriptions has gained research interest with the introduction of the DiDeMo
dataset~\cite{anne2017localizing}. The DiDeMo dataset provides open domain
videos with NL descriptions for temporal moment. Each NL description refers to a
specific temporal moment and describes the main event in the video. However,
DiDeMo does not provide spatial localization of the target within a frame and
videos collected are intended for event recognition instead of tracking.

The spatial-temporal localization by NL description task is first introduced by
Yamaguchi \etal~\cite{yamaguchi2017spatio}. The ActivityNet dataset is annotated
with NL descriptions to facilitate the training and evaluation of the proposed
task. However, the temporal retrieval in \cite{yamaguchi2017spatio} entails
retrieving the target video clip from a set of video clips. On the contrary, the
goal of the proposed vehicle tracking and retrieval task is to temporally
localize the target object within one sequence of video. Additionally, the
targets in the ActivityNet-NL take up most of the frame and cannot serve as a
tracking benchmark. In this paper, we focus on building a multi-target tracking
with NL description benchmark. The NL descriptions can be used to localize
targets both temporally and spatially in multi-view videos, providing benchmarks
for multiple Vision+NL tasks.

\begin{figure}
    \includegraphics[width=\columnwidth, page=1, trim=0 1cm 8cm 0, clip]{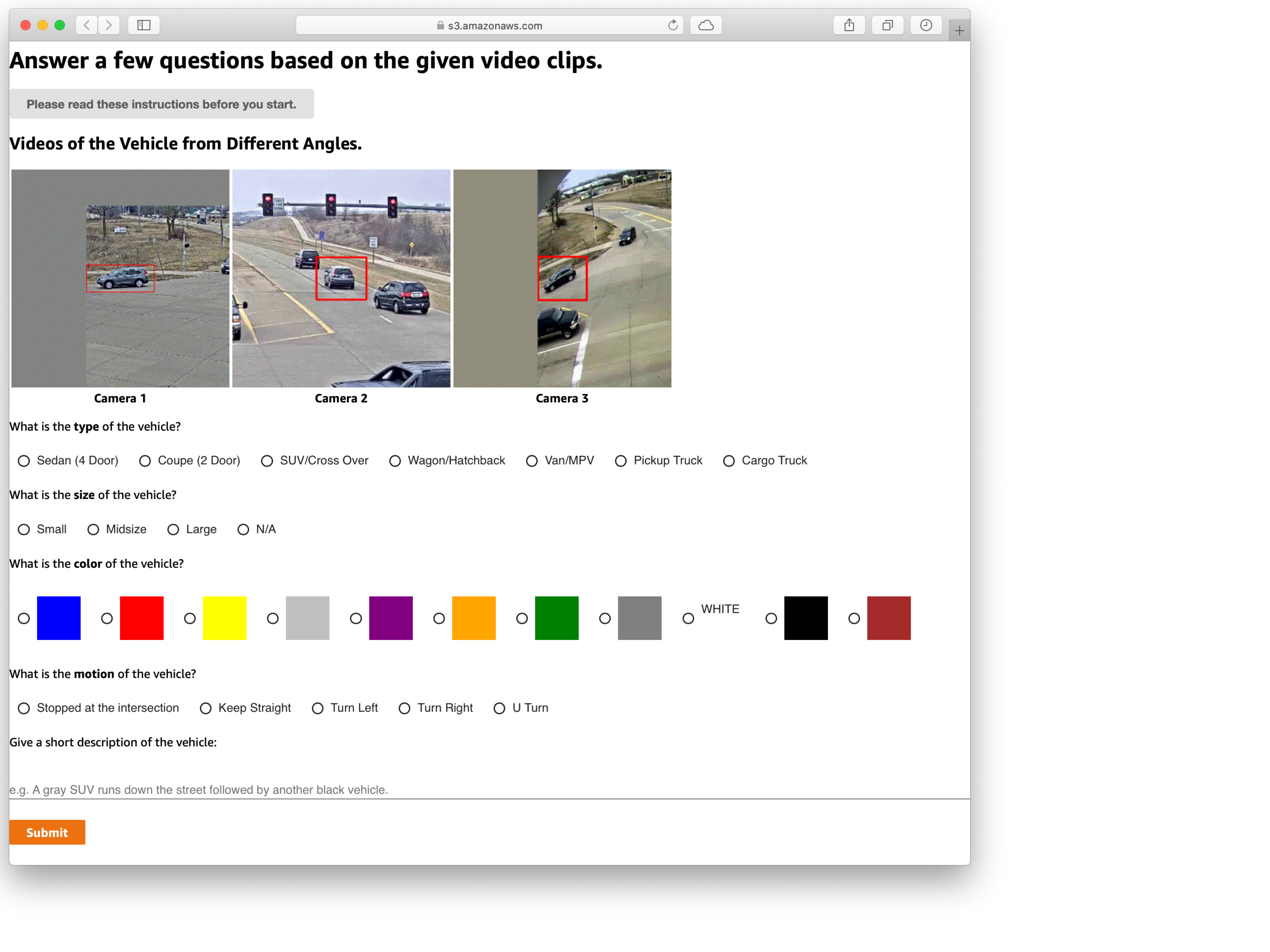}
  \caption{
    Screenshot of the platform we use to collect the CityFlow-NL dataset.
    Annotators are given detailed instructions on how to annotate NL
    descriptions for the vehicle targets presented in multi-view GIFs.
  }
  \label{fig-website}
  \vspace{-0.2 in}
\end{figure}

\begin{figure*}
  \includegraphics[width=\textwidth, page=1, trim=0 0 0 0, clip]{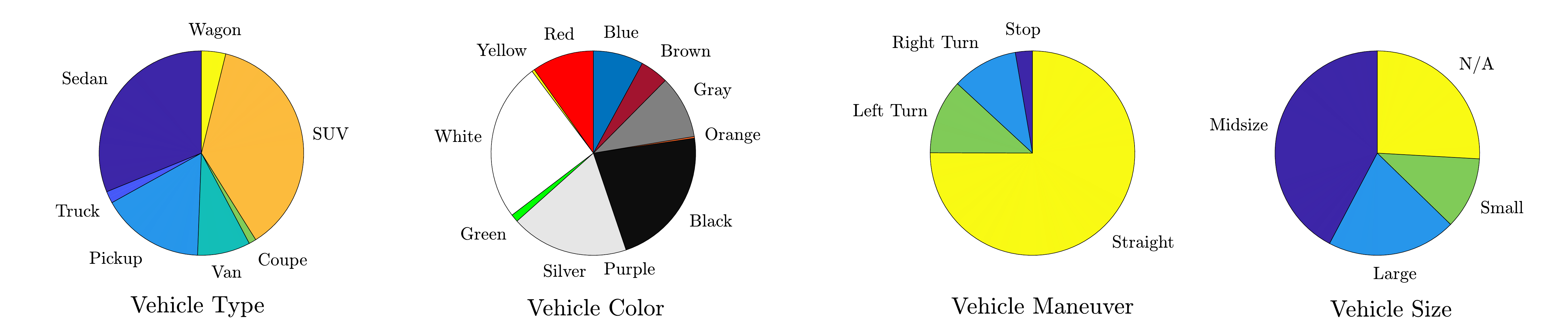}
    \caption{
    Distribution of vehicle type, color, maneuver, and size for the CityFlow-NL.
    A majority vote mechanism between three different annotators is implemented
    during the annotation process to automate the verification of the annotation
    and ensures the consensus and quality of the NL descriptions.
  }
  \label{fig-dist-attribute}
  \vspace{-0.2 in}
\end{figure*}

\section{CityFlow-NL Benchmark}
In this section, we describe the statistics of the proposed CityFlow-NL
benchmark and how we extend and annotate the CityFlow
Benchmark~\cite{tang2019cityflow}. 

\subsection{Overview}
We extend the CityFlow Benchmark with Natural Language (NL) Descriptions for
each target vehicle. The CityFlow-NL makes it possible to build and evaluate
systems that can jointly leverage both language and visual modalities. The
proposed CityFlow-NL consists of 666 targets vehicles in 3,028 (single-view)
tracks from 40 calibrated cameras, and 5,289 unique NL descriptions. The average
number of frames a target vehicle shows in the CityFlow Benchmark is 75.85. The
distribution of the number of frames of target vehicles is shown in
Fig.~\ref{fig-distribution}.

A web based platform, shown in Fig.~\ref{fig-website}, is designed to collect NL
annotations.  We provide multi-view video clips with red bounding boxes drawn
around each vehicle target and direct crowd-sourcing workers to give a detailed
NL description of the target vehicle that can be used to uniquely identify the
target vehicle from other vehicles. We collect annotations for each multi-view
track with at least three crowd-sourcing workers from Amazon SageMaker.

We split multi-view tracks from the CityFlow Benchmark by the timestamps of
cameras into multiple multi-view (or single-view) tracks that do not overlap
with each other temporally. For example, if a target shows up in camera 1 and
camera 2 in two temporal segments that do not overlap with each other, we
generate two separate tracks for our annotation process.  For each multi-view
track, we clip video segments from all views with bounding boxes drawn around
the target vehicle.

\begin{table}
    \begin{center}
        \begin{tabularx}{\columnwidth}{X|cccc}
            \hline
            Dataset & \# NL& T & MV & MT \\
            \hline
            CityFlow-NL & 5,289 & Yes & Yes & Yes\\
            LaSOT~\cite{fan2019lasot} & 1,400 & Yes & No & No\\
            OTB-99-LANG~\cite{li2017tracking} & 99 & Yes & No & No\\
            ActivityNet-NL~\cite{yamaguchi2017spatio} & 30,365 & No & No & No\\
            \hline
        \end{tabularx}
        \caption{
            Comparison between NL annotated computer vision datasets. T stands 
            for whether the dataset is designed for visual tracking; MV stands
            for multi-view (calibrated); MT stands for multi-target. The
            CityFlow-NL is the first MTMC for NL dataset and is the largest NL
            annotated tracking benchmark in terms of the number of NL.
        }
        \label{tbl-comparison}
        \vspace{-0.3 in}
    \end{center}
\end{table}

\subsection{Collecting Attributes of Target Vehicle}
Prior to a crowdsourcing worker generating an NL description (in the form of a declarative
sentence of an arbitrary length), the crowdsourcing worker is asked to tag the multi-view
video clip using predefined attributes. The distribution of these attributes are
summarized in Fig.~\ref{fig-dist-attribute}.

Performing such a preliminary task provides two benefits. First, it tends to
focus every crowdsourcing worker's attention on a common set of visual and linguistic
concepts while crowdsourcing worker remain free to create an NL description in any form.
Second, the attributes selected by crowdsourcing workers facilitate automatic verification
of an crowdsourcing worker's engagement with the task and consistency among the various
crowdsourcing workers of the same video clip. Since we direct crowdsourcing workers to
annotate NL descriptions in free form, it introduces uncertainty to the
annotation process and a direct verification of the annotated NL descriptions is
infeasible. Thus, a majority voting mechanism between annotations for the
attributes (color, motion, maneuver) from three different crowdsourcing workers
is used to automate the verification of the annotation and ensures the quality
of the NL descriptions. If no agreement between the three workers is reached,
the annotations are discarded and the target vehicle is re-annotated by another
three workers. 

As was the case with the CityFlow dataset~\cite{tang2019cityflow}, the
per-target attributes are not considered part of our CityFlow-NL dataset. One
reason is that the designers of the CityFlow Benchmark made a decision to hide
attributes when releasing their challenge dataset for the AI City
Challenge~\cite{naphade2017nvidia}, in part to prevent challenge participants
from gaining unfair advantage. Another reason is that while attribute-based
tracking received attention in the past~\cite{tripathi2019tracking}, the
motivation for using attributes is not as apparent given strong NL models, and
the requirements of real-world scenarios. Indeed, as shown
in~\cite{feng2020real} NL-based tracking decisively outperforms
attribute-based tracking on a challenging benchmark.  The attributes are not
equivalent to the NL descriptions, where additional information including
vehicle maneuver and relations to other targets are essential to a variety of
problems.

\subsection{Collecting NL Descriptions}
NL descriptions of the same target can be very different, as NL descriptions are
subjective and may focus on different aspects of the target. Examples of such
annotations are shown in Fig.~\ref{fig-nl-annotations}. The NL descriptions we
collect tend to describe the vehicle color, the vehicle maneuver, the traffic
scene, and the relation of the vehicle with other vehicles in the scene. We
specifically direct crowdsourcing workers to describe the intrinsic motion for
all maneuvers in the multi-view video clip. Moreover, some conflicting
descriptions are acceptable due to color distortions and view-angle differences
in these multi-view videos. Our majority vote mechanism identifies consensus
among annotators, given the natural variation in their NL descriptions.

\subsection{Comparison to Other NL+Vision Datasets}
We compare the CityFlow-NL to other NL annotated vision datasets and presented
in Fig.~\ref{fig-star} and Tbl.~\ref{tbl-comparison}. The CityFlow-NL is the
first dataset built for the MTMC by NL task.
Comparing to prior NL annotated video datasets, the proposed CityFlow-NL focuses
on building a dataset for multi-object tracking and keeps the unique challenges
for visual tracking as we discussed in Sec.~\ref{sec-related-works}. Each of the
NL descriptions in our dataset describes a specific target for the entire
duration in the multi-view video. Our annotation process avoids potential
observation and annotation biases from crowdsourcing workers for vehicle motion
by presenting multi-view GIFs, using multiple-choice questions for attributes
and a majority vote mechanism implemented during the annotation process.

\section{Vehicle Retrieval by NL Task}
We now define the Vehicle Retrieval by Natural Language Description Task and
present how we use the proposed CityFlow-NL Benchmark for it. We consider this
task as an essential first step towards tackling more advanced Vision+NL tasks
like the MTMC by NL description task.

\subsection{Task Definition}
For the purpose of the retrieval by NL task, we utilize the proposed CityFlow-NL
Benchmark in a \textit{single-view} setup, although the CityFlow-NL could be
used for retrieval tasks with multi-view tracks. For each single-view vehicle
track, we bundle it with a query that consists of three different NL
descriptions for training. During testing, the goal is to retrieve and rank
vehicles tracks based on NL queries.

This variation of the proposed CityFlow-NL contains 2,498 tracks of vehicles
with three unique NL descriptions each. Additionally, 530 unique vehicle tracks
together with 530 query sets (each annotated with three NL descriptions) are
curated for testing. 

The proposed NL tracked-object retrieval task offers unique challenges versus
action recognition tasks and content-based image retrieval tasks. In particular,
different from prior content-based image retrieval
systems~\cite{guo2018dialog,hu2016natural,mao2016generation}, retrieval models
for this task need to consider the relation contexts of a vehicle track and
motion within the track as well. While the action recognition by NL description
task~\cite{anne2017localizing} temporally localizes a moment within a video, the
proposed tracked-object retrieval task requires both temporal and spatial
localization within a video.

\subsection{Evaluation Metrics}
The Vehicle Retrieval by NL Description task is evaluated using standard metrics
for retrieval tasks~\cite{manning2008introduction}.  We use the Mean Reciprocal
Rank (MRR) as the main evaluation metric. Recall @ 5, Recall @ 10, and Recall @
25 are also evaluated for all models.

For each query in the testing split, we rank all 530 candidate tracks using the
retrieval system and evaluate the retrieval performance based on the
above-mentioned metrics. 

\subsection{The Retrieval Baseline Model}
\label{sec-retrieval-baseline}

\begin{figure}
  \includegraphics[width=\columnwidth, page=5, trim=0 10.5cm 1cm 0, clip]{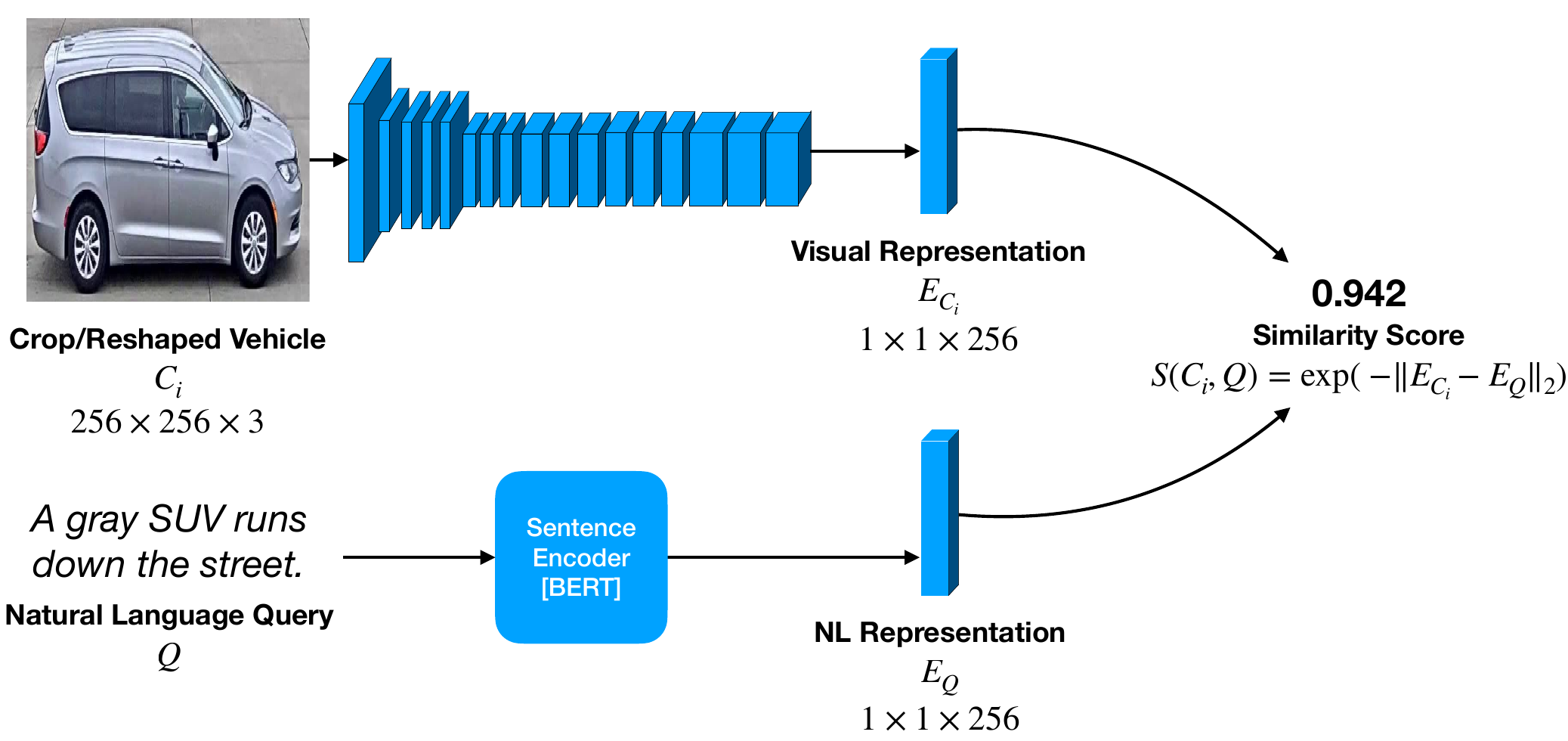}
  \caption{
    The baseline retrieval model inspired by content-based image retrieval
    models~\cite{guo2018dialog,hu2016natural,mao2016generation}. A similarity
    between the NL representation of the NL query and the visual representation
    of the input frame crop is measured by the Siamese network.
  }
  \label{fig-baseline-model}
  \vspace{-0.2 in}
\end{figure}

\begin{figure*}
    \begin{subfigure}{\textwidth}
        \includegraphics[width=\columnwidth, page=10, trim=0 23cm 0cm 0, clip]{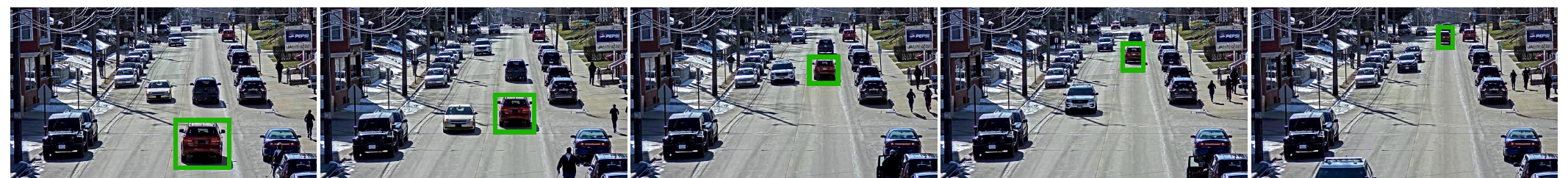}
        \caption{A target vehicle track for retrieval. The NL descriptions given
        are: \textit{``A red wagon goes straight'', ``Red SUV going straight in
        the right lane.''} and \textit{``A red SUV head straight down the
        road.''} The baseline retrieval model gives this track the average
        similarity score at 0.81 and ranks this track the fifth. Notice that the second 
        NL description specified that the target is moving in the right lane.}
    \end{subfigure}
    \begin{subfigure}{\textwidth}
        \includegraphics[width=\columnwidth, page=10, trim=0 16.5cm 0cm 6cm, clip]{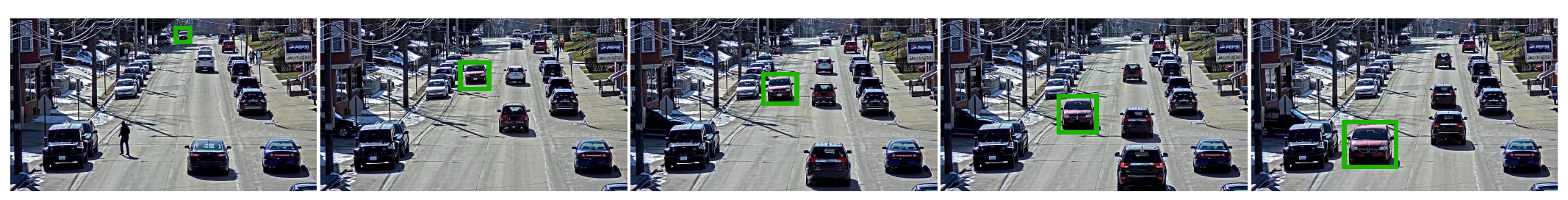}
        \caption{The baseline retrieval model ranks this track the first with
        the highest average similarity score at 0.84. This vehicle shares the same 
        size, type, color, and motion pattern with the target vehicle for retrieval. 
        The baseline retrieval model fails to capture the spatial referring
        expression (\textit{"in the right lane''}) from the second NL description.
        }
    \end{subfigure}
    \caption{
      Example of a failed retrieval by the baseline retrieval model.
    }
    \label{fig-baseline-results}
    \vspace{-0.2 in}
  \end{figure*}

We build the baseline model to measure the similarities between a vehicle track
$\mathcal{T}$ and a NL description $Q$. Our baseline formulation is inspired by
content-based image retrieval systems~\cite{young2014image} and image captioning
models~\cite{johnson2016densecap,vinyals2015show}. The overview of the baseline
model is shown in Fig.~\ref{fig-baseline-model}.

The vehicle track is defined as a sequence of frames in a video clip.
$\mathcal{T} = \left\{F_1,\cdots,F_t\right\}.$ We use the ground truth bounding
box of the target vehicle to crop each frame and reshape the sequence of crops,
denoted as $C$, to the same shape of $(256, 256)$. This sequence of reshaped
frame crops of the target, $\left\{C_1, \cdots, C_t\right\}$,  is used as the
visual representation of the vehicle. We build a Siamese embedding model for the
retrieval task, where each crop $C_i$ is embedded by a
ResNet-50~\cite{he2016identity} pretrained on
ImageNet~\cite{russakovsky2015imagenet}, denoted as $E_{C_i}$. We use a
pretrained BERT~\cite{devlin2018bert} to embed the NL description $Q$ into a 256
dimensional vector, denoted as $E_Q$.

The similarity between $C_i$ and $Q$ is defined as:
\begin{equation}
\mathcal{S}\left({C_i}, Q\right) = \exp\left(- \|E_{C_i} - E_Q\|_2\right).
\end{equation}

For the baseline retrieval model, both positive and negative pairs of vehicle
image crop $C_i$ and $Q$ are constructed for training.  Negative pairs of $C_i$
and $Q$ are built by randomly selecting NL descriptions that describe other
targets. The baseline retrieval model is trained with a cross entropy loss and
an initial learning rate of 0.001 for 20 epochs on 2 GPUs using a stochastic
gradient descent optimizer. 

For the purpose of inference, we measure the similarity between a test track and
a test query as the average of the similarities between all pairs of crops in
the track and NL description in the test query. \ie
\begin{equation}
\mathcal{S}\left(\mathcal{T}, \{Q_1, Q_2, Q_3\}\right) =
  \frac{1}{3\cdot t}\sum_{j=1}^{3}\sum_{i=1}^{t}\mathcal{S}(C_i, Q_j).
\end{equation}

\subsection{Retrieval Results and Analysis}
For each test query, we compute this similarity on all test tracks and rank the
tracks for evaluation. The baseline model achieves an MRR of 0.0269, Recall @ 5
of 0.0264, Recall @ 10 of 0.0491, Recall @ 25 of 0.1113.

An example of a failure case from the baseline retrieval model is shown in
Fig.~\ref{fig-baseline-results}. In this specific case, positional referring
expressions (``in the right lane'') is given in the NL description, but the
baseline retrieval model does not consider the position of the target within
the frame. Moreover, it is worth noting that the baseline model does not
consider the motion pattern nor the context of vehicle tracks explicitly and the
retrieval performance could be further improved by measuring the motion within
the frame. 

\begin{figure}
    \includegraphics[width=\columnwidth, page=7, trim=0 10.5cm 5cm 0, clip]{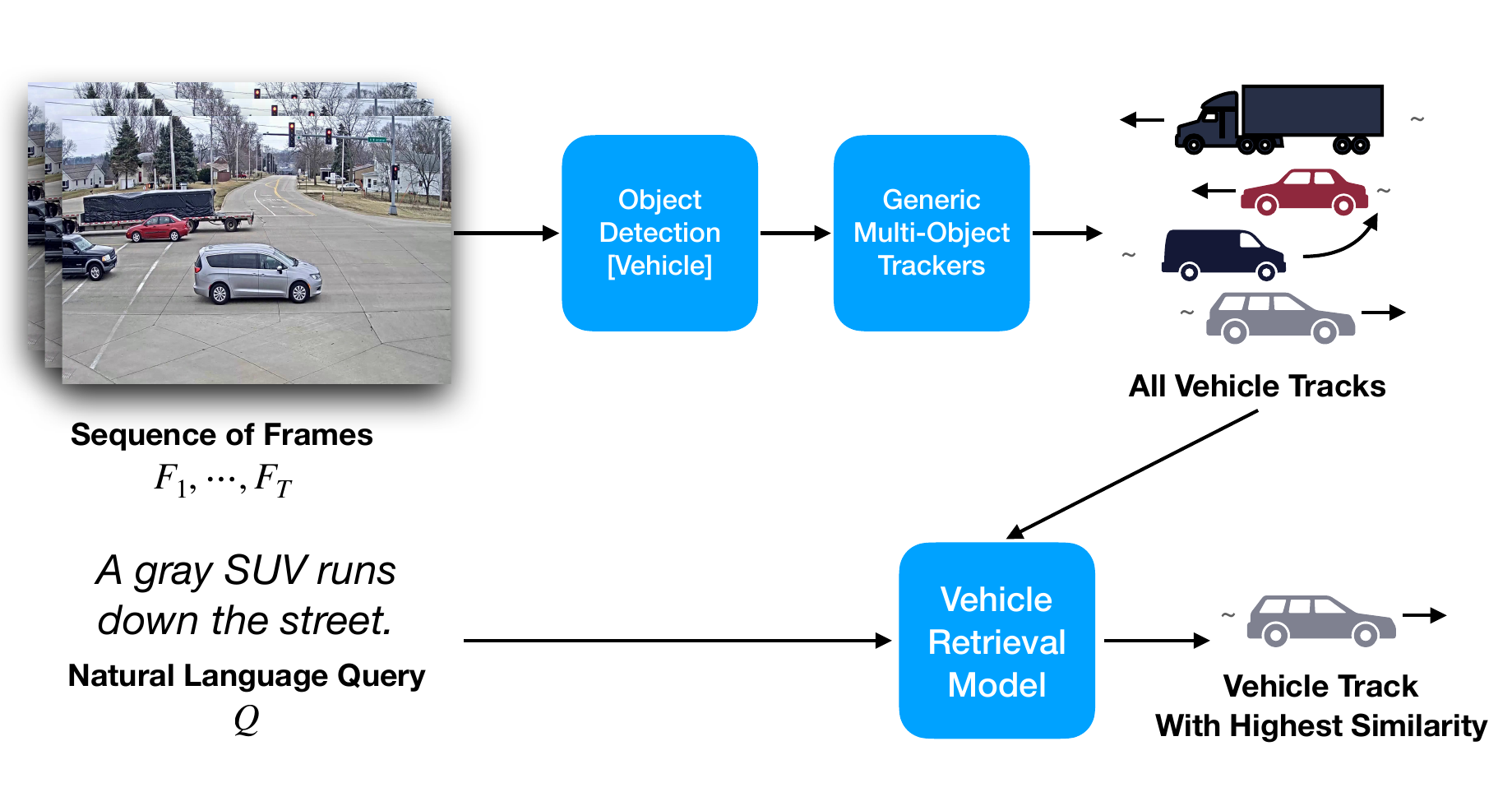}
  \caption{
    The baseline tracking model comprises three steps. The first step is to
    detect all vehicles in every frame. The second step is to utilize a generic
    multi-object tracker to obtain all vehicle tracks. In the last step, the
    baseline retrieval model is used to rank predicted tracks by their
    similarities and predict the target vehicle track.
  }
  \label{fig-baseline-tracking-model}
  \vspace{-0.2 in}
\end{figure}

\section{Vehicle Tracking by NL Task}
We now describe how we use the proposed CityFlow-NL Benchmark for the vehicle
tracking by NL description task.

\subsection{Problem Definition}
We further investigate the potential of the proposed retrieval model and extend
it to the tracking by NL task. With the same query as in the retrieval task,
together with a sequence of frames $(F_1, \cdots, F_T)$ as inputs, the goal is
to track the specific target the NL query refers to in the form of a sequence of
bounding boxes. For each frame, the tracker should either predict the bounding
box for the target or predict that the target is not present in the frame.

\subsection{Evaluation Metrics}
The vehicle tracking by NL description task is evaluated, following popular
object tracking protocols, using the Area Under Curve (AUC) of the Success Rate
vs. Intersection over Union (IoU) thresholds and Normalized
Precision~\cite{wu2013online}.

The success is defined as 0 for frames where the prediction is ``not present''
while the ground truth is available, or the ground truth is ``not present'' and
a prediction is made. For frames with both a ground truth bounding box and a
prediction, the success is measured by computing the IoU between the ground
truth bounding box and the prediction and comparing this against a predefined
threshold. The AUC of the success rate at different IoU thresholds is reported
as the performance for each tracker~\cite{wu2013online}.

\subsection{The Tracking Baseline Model}
Intuitively, to perform tracking of the target vehicle using NL descriptions, we
use our baseline retrieval model introduced in Sec.~\ref{sec-retrieval-baseline}
for inference on tracks predicted by multi-object trackers. The baseline model
is shown in Fig.~\ref{fig-baseline-tracking-model}.

The tracking baseline model consists of three stages. The first stage of the
model is an object detector which localizes all vehicles in every frame. In the
second stage, we use a generic multi-object tracker to predict vehicle tracks on
the test split of the CityFlow-NL.  In the last stage, we use the baseline
retrieval model introduced in Sec.~\ref{sec-retrieval-baseline} to rank the
predicted vehicle tracks for each query. The vehicle track with the highest
average similarity score is returned by the baseline tracker as the prediction.

The baseline vehicle tracking model first localizes and tracks all vehicles in a
video and predicts a retrieval score for each vehicle track in the second pass.
The time complexity of retrieval baselines model is $O(M\times N)$, where $M$ is
the size of the query set and $N$ is the number of tracks predicted by the
multi-object tracker.

In our experiments, we tested our baseline tracker with three different
multi-object trackers: DeepSORT~\cite{wojke2017simple},
MOANA~\cite{tang2019moana}, and TNT~\cite{hsu2019multi} with three different
sets of vehicle detections: MaskRCNN~\cite{he2017mask},
YOLO3~\cite{redmon2018yolov3}, and SSD512~\cite{liu2016ssd}. We use the same
model we trained for the retrieval task for the tracking task. Results are
summarized in Tbl.~\ref{tbl-tracking-results}. Trackers using SSD detections are
outperforming others with a higher recall rate. Theses baseline trackers are
competitive with each other.

\begin{figure}
    \includegraphics[width=\columnwidth, page=6, trim=0 10.5cm 0cm 0, clip]{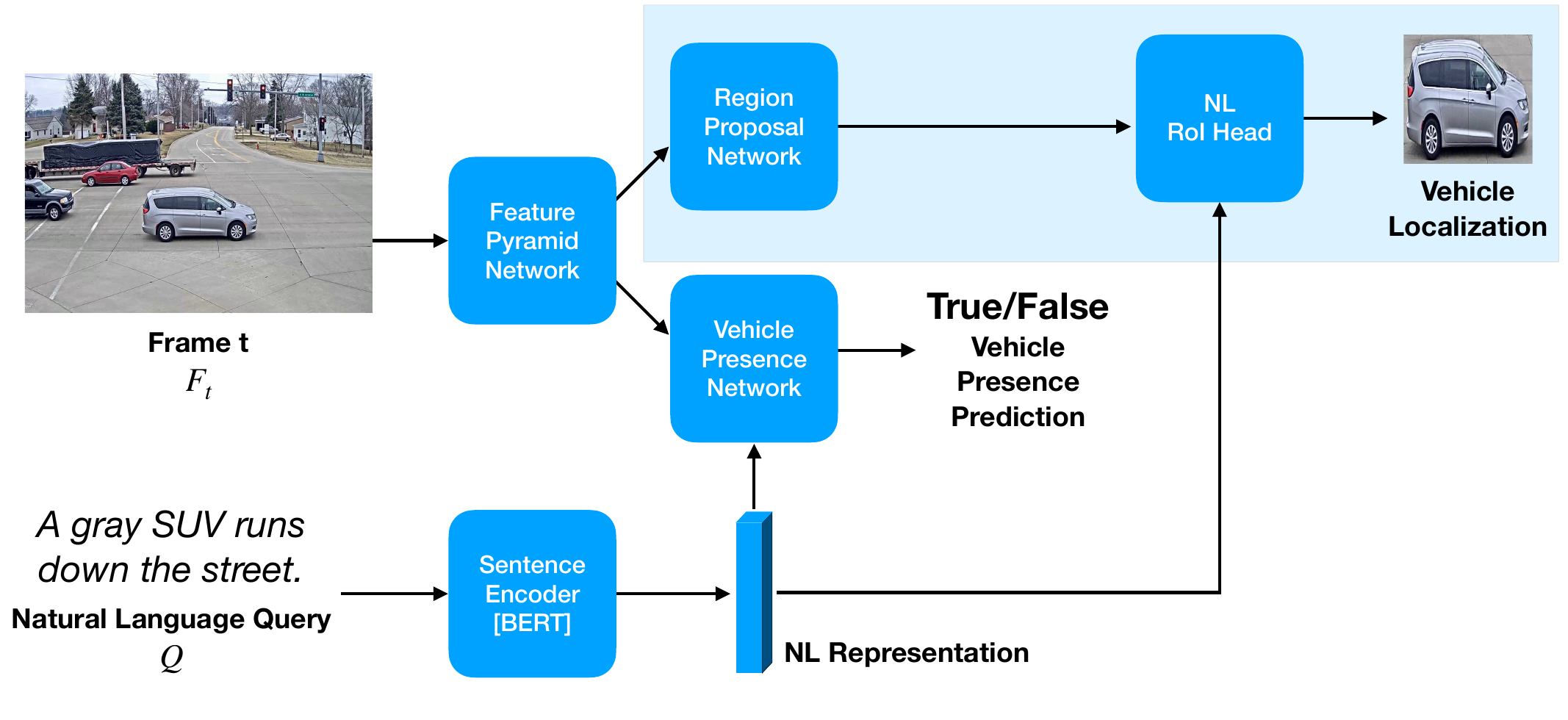}
  \caption{
    The proposed Vehicle Tracking Network (VTN) has two branches: an NL
    based Vehicle Presence branch and a NL based vehicle localization branch.
    The vehicle presence branch predicts whether the vehicle target of the given
    NL query is present in the given frame. The vehicle localization branch
    makes region proposals and predicts bounding boxes based on the NL query.
  }
  \label{fig-vtrm}
  \vspace{-0.2 in}
\end{figure}

\subsection{Vehicle Tracking Network (VTN)}
Instead of the two-stage multi-object tracking and NL retrieval approach, we now
introduce a more efficient and accurate Vehicle Tracking Network (VTN), which is
summarized in Fig.~\ref{fig-vtrm}. Similar to prior works on single object
tracking with NL description~\cite{feng2020real,feng2019robust,li2017tracking},
we design the VTN to be a one-shot tracker by detection that is adapted to
utilize an NL description. The VTN first extracts the visual features of a
frames $F_t$ using a Feature Pyramid Network (FPN)~\cite{lin2017feature}. The
visual features are then used in two separate branches: a Vehicle Presence
Network (VPN) and a Vehicle Localization Network (VLN).

\begin{figure}
    \includegraphics[width=\columnwidth, page=8, trim=0 3cm 7cm 0, clip]{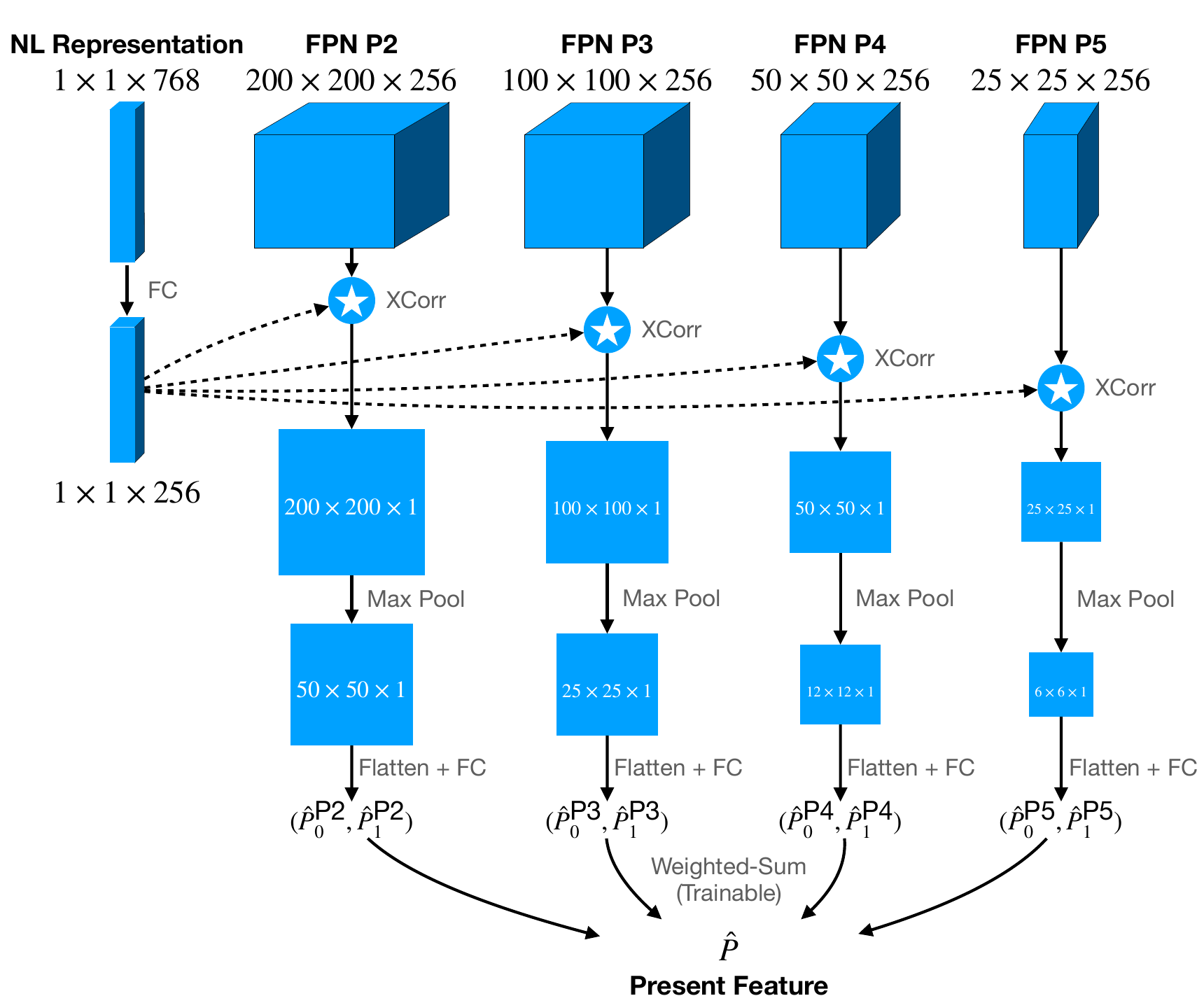}
  \caption{
    The Vehicle Presence Network (VPN), part of the proposed VTN, as shown in Fig.~\ref{fig-vtrm}.
  }
  \label{fig-vpn}
  \vspace{-0.2 in}
\end{figure}

The VPN, shown in Fig.~\ref{fig-vpn}, indicates the target's presence in the
given frame to initiate or terminate a track. The vehicle tracking by NL
description task differs from single object tracking by NL description
task~\cite{li2017tracking} as the presence of the target is not guaranteed for
most of the frames. Thus, a cross-correlation (XCorr)
operation~\cite{bertinetto2016fully} between the NL representation and the
visual features from the four stages of the FPN (``P2'' thru ``P5'') is used to
localize the target temporally. A weighted average trained between the four
presence features is then used to compute the output presence feature, denoted
as $P$. A softmax function between the two elements in $P$ is used to determine
the presence probability $\hat{P}$ of the vehicle target based on the NL query,
\ie
\begin{equation}
  \hat{P} = \frac{e^{P_1}}{e^{P_0} + e^{P_1}}.
\end{equation}

We use $P^*$ to denote the ground truth presence of the target, where $P^* = 1$
represents the presence and $P^* = 0$ represents that the target is not present
in the frame. A binary cross entropy loss is used to train the presence branch:
\begin{equation}
  \Loss_\text{Presence} = - \hat{P}\cdot \log P^* - (1-\hat{P})\cdot \log (1- P^*).
\end{equation}

\begin{table*}
    \begin{center}
    \begin{tabular}{c|cc|cc}
        \hline
        \multicolumn{3}{c|}{Model} & AUC Success (\%) & Normalized Precision (\% at 0.5) \\ 
        \hline
        \multirow{9}{2cm}{\centering Baseline Retrieval Model} & \multirow{3}{*}{DeepSORT~\cite{wojke2017simple}} & Mask RCNN~\cite{he2017mask}& 0.66 & 1.11 \\
        
        & & YOLOv3~\cite{redmon2018yolov3} & 0.92 & 1.78 \\
        & & SSD512~\cite{liu2016ssd} & \textit{2.13} & 1.68\\
        \cline{2-5}
        & \multirow{3}{*}{MOANA~\cite{tang2019moana}} & Mask RCNN~\cite{he2017mask}& 0.52 & 0.52\\
        & & YOLOv3~\cite{redmon2018yolov3} & 1.69 & 1.80\\
        & & SSD512~\cite{liu2016ssd}  &0.94 & 1.61\\
        \cline{2-5}
        & \multirow{3}{*}{TNT~\cite{tang2019moana}} & Mask RCNN~\cite{he2017mask}& 1.07 & 1.55\\
        & & YOLOv3~\cite{redmon2018yolov3} & 0.66 & 1.06\\
        & & SSD512~\cite{liu2016ssd} & 1.56 & \textit{2.05}\\
        \hline
        \multicolumn{3}{c|}{Vehicle Tracking Network (VTN)} & \textbf{5.93} & \textbf{3.79}\\
        \hline
    \end{tabular}
    \caption{AUC of Success Rate Plot and Normalized Precision on the test split
    of the CityFlow-NL. We compare the VTN (Fig.~\ref{fig-vtrm}) against the
    baseline "track-then-retrieve" approach
    (Fig.~\ref{fig-baseline-tracking-model}) that was instantiated with baseline
    multi-object trackers. DeepSORT~\cite{wojke2017simple},
    MOANA~\cite{tang2019moana}, and TNT~\cite{hsu2019multi} with three different
    sets of vehicle detections: MaskRCNN~\cite{he2017mask},
    YOLO3~\cite{redmon2018yolov3}, and SSD512~\cite{liu2016ssd} are reported
    here. Best and second best entries are marked with bold and italic fonts respectively.} 
    \label{tbl-tracking-results}
    \vspace{-0.2in}
\end{center}
\end{table*}

\begin{figure*}
    \includegraphics[width=\textwidth, page=9, trim=0 23cm 0cm 0, clip]{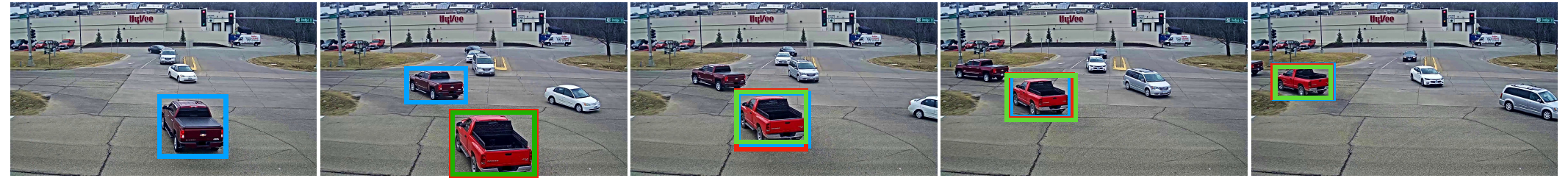}
    \caption{
      Failure case of the VTN when compared to the baseline tracking model. Blue
      bounding boxes are results from the VTN; red bounding boxes are results
      from the baseline tracking model with DeepSORT~\cite{wojke2017simple} and
      YOLOv3~\cite{redmon2018yolov3}; green bounding boxes are the ground truth.
      It is worth notice that the VTN is an one pass and online tracker, which
      predicts the highest scored bounding box when the VPN predicts that the
      target is present. Thus, the false positive rate is higher than the
      baseline tracking model, especially when there are contrasting vehicle
      targets in the same video.
    }
    \label{fig-vtn-results}
    \vspace{-0.2 in}
\end{figure*}

The VLN is a modified version of the Faster-RCNN~\cite{ren2015faster}, which
proposes regions and localizes the vehicle target based on the NL query. We use
a standard Region Proposal Network (RPN) with anchor sizes of 32, 64, 128, 256,
512 and anchor ratios of 0.5, 1.0 and 2.0 for generating a set of region
proposals. The NL Region of Interest (RoI) Head not only predicts the regression
and classification as in a standard object detection RoI head, it also measures
similarities between visual features for each region proposals and the NL
representation, denoted as $\hat{S}$. The NL similarity layers are trained with
a cross entropy loss that is similar to the classification loss of the RoI loss:
\begin{equation}
  \Loss_\text{NL} = \sum_\text{all regions} - \hat{S} \log S^* - (1- \hat{S}) \log (1-S^*).
\end{equation}
where $S^*$ is the ground truth similarity. When curating the ground truth
labels for training the RPN and NL RoI Head, all vehicles in the frame are
marked as positive. However, the target similarity $S^*=1$ if and only if the
region's target regression ground truth bounding box is exactly what the NL
description refers to.

Thus, the loss for the localization branch is:
\begin{equation}
  \Loss_\text{Localization} = \Loss_\text{RPN} + \Loss_\text{RoI} + \Loss_\text{NL},
  \label{eq-loss-rpn}
\end{equation}
where $\Loss_\text{RPN}$ and $\Loss_\text{RoI}$ is the same as
in~\cite{ren2015faster}. The two branches in the VTN are jointly trained
end-to-end with the above-mentioned losses:
\begin{equation}
  \Loss_\text{VTN} = \Loss_\text{Presence} + \Loss_\text{Localization}.
\end{equation}

The VTN is initialized with pretrained weights of a Faster-RCNN with FPN trained
for MSCOCO. We train the RPN and RoI Head with an initial learning rate of 0.001
and the VPN with an initial learning rate of 0.01. The learning rate is decayed
by 0.5 every 2,000 steps with a batch size of 16 frames distributed across 4
GPUs using a stochastic gradient descent optimizer.

When used during inference, we use the VPN with a threshold of 0.5 that is
validated on the training split to indicate if the target is present in a frame.
If the target is found in the frame, the VLN is then used to predict the
similarities between detected vehicle bounding boxes and the NL query. We apply
a sub-window attention~\cite{feng2019robust} to the similarities and pick the
highest scored one as the prediction for the frame.

Results for the vehicle tracking task are presented in
Tbl.~\ref{tbl-tracking-results}. The VTN outperforms baseline trackers and runs
at 20 fps on a single GPU. As the VTN performs the NL retrieval at an earlier
step during the detection phase, the recall of the targets is improved compared
to the baseline trackers, which perform the NL retrieval at the last step after
multi-object tracking. A failure case of the VTN is shown in
Fig.~\ref{fig-vtn-results}. As the VTN is designed as an online tracker by NL
detection and predicts the highest scored bounding box when the VPN predicts
that the target is present, the false positive rate is higher than the baseline
tracking model, especially when contrasting vehicle targets are present.

\section{Conclusion}
This paper has presented CityFlow-NL, the first city-scale multi-target
multi-camera tracking with NL descriptions dataset that provides precise NL
descriptions for multi-view ground truth vehicle tracks. Our NL annotations can
be used to benchmark tasks like vehicle tracking and retrieval by NL, motion
pattern analysis with NL, and MTMC with NL descriptions. We introduce two basic
benchmark tasks with multiple baseline models utilizing the proposed
CityFlow-NL: The Vehicle Retrieval by NL task and The Vehicle Tracking by NL
task, facilitating future research on more advanced Vision+NL tasks and systems.

{\small
\bibliographystyle{ieee}
\bibliography{bib}
}
\end{document}